\documentclass[letterpaper]{article} 
\usepackage{aaai24}  
\usepackage{times}  
\usepackage{helvet}  
\usepackage{courier}  
\usepackage[hyphens]{url}  
\usepackage{graphicx} 
\usepackage{natbib}  
\usepackage{caption} 
\usepackage{algorithm}
\usepackage{algorithmic}

\usepackage{booktabs}
\usepackage{multirow}
\usepackage{diagbox}
\usepackage{amsthm}
\usepackage{amssymb}
\usepackage{amsmath}
\usepackage{units}
\newtheorem{theorem}{Theorem}
\newtheorem{definition}{Definition}

\usepackage{newfloat}
\usepackage{listings}
\DeclareCaptionStyle{ruled}{labelfont=normalfont,labelsep=colon,strut=off} 
\floatstyle{ruled}
\newfloat{listing}{tb}{lst}{}
\floatname{listing}{Listing}
\title{Self-Supervised Representation Learning with Meta Comprehensive Regularization}

\author {
    Huijie Guo\textsuperscript{\rm 1}$^{,}$\equalcontrib,
    Ying Ba\textsuperscript{\rm 4}$^{,}$\textsuperscript{\rm 5}$^{,}$\equalcontrib,
    Jie Hu\textsuperscript{\rm 6},
    Lingyu Si\textsuperscript{\rm 2}$^{,}$\textsuperscript{\rm 3},
    Wenwen Qiang\textsuperscript{\rm 2}$^{,}$\textsuperscript{\rm 3}$^{,}$\thanks{Corresponding author.},
    Lei Shi\textsuperscript{\rm 1}$^{,}$\thanks{Co-corresponding author.}
}
\affiliations {
    \textsuperscript{\rm 1}Beihang University\\
    \textsuperscript{\rm 2}Institute of Software Chinese Academy of Sciences\\
    \textsuperscript{\rm 3}University of Chinese Academy of Sciences\\
    \textsuperscript{\rm 4}Gaoling School of Artificial Intelligence, Renmin University of China\\
    \textsuperscript{\rm 5}Beijing Key Laboratory of Big Data Management and Analysis Methods\\
    \textsuperscript{\rm 6}Meituan\\
    \{guo\_hj, leishi\}@buaa.edu.cn, yingba@ruc.edu.cn, hujie@ios.ac.cn, \{lingyu, qiangwenwen\}@iscas.ac.cn
}

\usepackage{bibentry}

\begin{document}

\maketitle

\begin{abstract}
    Self-Supervised Learning (SSL) methods harness the concept of semantic invariance by utilizing data augmentation strategies to produce similar representations for different deformations of the same input. Essentially, the model captures the shared information among multiple augmented views of samples, while disregarding the non-shared information that may be beneficial for downstream tasks. To address this issue, we introduce a module called CompMod with Meta Comprehensive Regularization (MCR), embedded into existing self-supervised frameworks, to make the learned representations more comprehensive. Specifically, we update our proposed model through a bi-level optimization mechanism, enabling it to capture comprehensive features. Additionally, guided by the constrained extraction of features using maximum entropy coding, the self-supervised learning model learns more comprehensive features on top of learning consistent features. In addition, we provide theoretical support for our proposed method from information theory and causal counterfactual perspective. Experimental results show that our method achieves significant improvement in classification, object detection and instance segmentation tasks on multiple benchmark datasets.
\end{abstract}

\section{Introduction}


Deep learning models have exhibited remarkable capabilities, leading to the widespread adoption of machine learning across diverse fields. Despite the impressive performance of supervised learning methods, their heavy reliance on labeled data for model training poses limitations on their generalization ability and scalability. To address this challenge, Self-Supervised Learning (SSL) has emerged as a promising paradigm that bridges the gap between supervised and unsupervised learning by generating supervised signals directly from the samples without the need for manual annotation. Currently, SSL has achieved remarkable results in computer vision ~\cite{tian2020contrastive,chen2021empirical,caron2020unsupervised} and natural language processing ~\cite{baevski2020wav2vec,akbari2021vatt,zhou2020pre}. 

The general framework of self-supervised representation learning consists of two key components: data augmentation and loss function, which try to learn invariance to the transformation generated by data augmentation on the same sample while maintain discrimination to different samples. In practice, data augmentation generates two augmented views of the same image by applying random strategies, such as Cutout ~\cite{devries2017improved}, Coloring~\cite{zhang2016colorful}, Random Cropping~\cite{takahashi2019data}, etc. Several studies~\cite{zheng2021ressl, shorten2019survey, zhang2022rethinking, tian2020makes} also have suggested that not all data augmentations are beneficial for downstream tasks. For instance, rotation invariance may help some flower categories but harm animal recognition ~\cite{xiao2020should}. Similarly, color invariance may have opposite effects on animal and flower classification tasks. Therefore, recent works have proposed adaptive augmentation strategies to adapt to different data and task environments~\cite{li2022metaug,yang2022identity}.

\begin{figure}[t]
    \centering
    \includegraphics[width=0.95\linewidth]{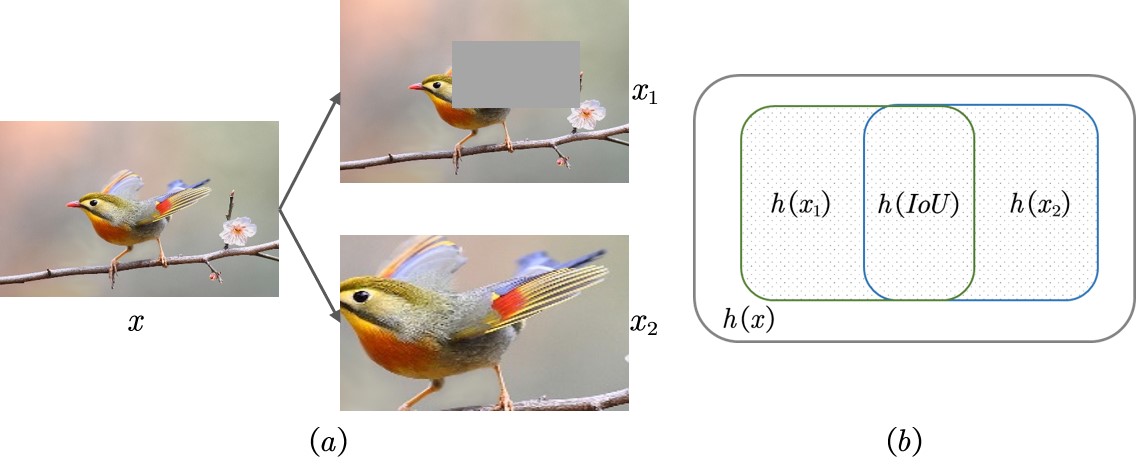}
    \caption{
    Loss of task-related information caused by data augmentation in SSL methods. ($a$), the positive sample pair $(x_a,x_b)$ can be obtained from the input $x$ by Random Cropping and Cutout. ($b$) formally presents the semantics related to label in different augmented views, where $h(\cdot)$ represents the amount of attributes related to the label in sample.
    }
    \label{fig:intro_1}
\end{figure}

Data augmentation strategies are widely used in SSL to create positive pairs of images that share the same label. However, these strategies may not preserve all the semantic information that is relevant to the label in the augmented views. For example, suppose an image’s label is “bird” and it only refers to the foreground object, not the background. Figure \ref{fig:intro_1}(a) shows two views of the same image $x$ created by Random Cropping and Cutout, denoted as $x_1$ and $x_2$. Note that $x_1$ contains the bird’s beak while $x_2$ does not, and $x_2$ contains the bird’s wings while $x_1$ does not. A common assumption in SSL is that the semantic content of an image should be invariant to the applied transformations. However, this assumption can be broken by the transformation methods and may not hold for all label-related attributes, such as the bird’s beak and wings. Figure \ref{fig:intro_1}(b) illustrates a function $h(\cdot)$ that measures the amount of label-related attributes in an image. The representations learned by SSL methods are based on the shared information between different augmented views, such as the Intersection over Union (IoU). However, this shared information may not capture the entire foreground of the input, and some label-related attributes may be dropped in the model training process. The more label-related information is preserved in the training process, the better the model can learn. Therefore, models trained using traditional SSL methods may exhibit subpar performance in downstream tasks due to the loss of label-related information during the training process.

To address the aforementioned issue, we propose utilizing a more comprehensive representation to guide the training of SSL model, enabling the model to focus on non-shared semantic information that might be beneficial for downstream tasks, thereby enhancing model's generalization capability. We propose a plug-and-play module called CompMod with Meta Comprehensive Regularization to guide the learning of SSL methods by obtaining comprehensive features. Specifically, we employ semantic complementarity to fuse augmented features in a low-dimensional space, utilizing a bi-level optimization mechanism to obtain comprehensive representation that guide the learning of SSL methods. Our contributions are the following:

\begin{itemize}

    \item From the information theory, we analyze that data augmentation in SSL may lead to the lack of task-related information, which in turn reduces the generalization ability of the model.
    
    \item We design a plug-and-play module, called CompMod, to induce existing SSL methods to learn comprehensive feature representations. CompMod ensures comprehensive feature exploration through a bi-level optimization mechanism and constrained extraction of features with maximum entropy coding, guaranteeing complete mining of feature completeness.

    \item A causal counterfactual analysis provides theoretical support for our proposed method. Empirical evaluations of the proposed method substantiate its superior performance in classification, object detection and instance segmentation tasks.
\end{itemize}

\section{Related Work}
\label{re_work}

Recently, various frameworks have emerged for self-supervised representation learning, which can be broadly classified into two types~\cite{garrido2022duality, balestriero2022contrastive}: sample-based and dimension-based contrastive learning methods.

Sample-based contrastive methods learn visual representations by constructing pairs of samples and applying contrastive loss function. These methods encourage the embeddings of augmented views of the same image to be close to each other, while simultaneously pushing away the embeddings of different images. Some notable methods, such as SimCLR~\cite{chen2020simple}, utilize InfoNCE as the loss function and rely on the quality and quantity of negative samples. However, these methods also necessitate greater computational resources. MoCo~\cite{he2020momentum} tackles this issue by constructing a dynamic dictionary bank that expands the pool of available negative samples. On the other hand, some studies have investigated whether SSL can still work without negative samples. BYOL~\cite{grill2020bootstrap} and SimSiam~\cite{chen2021exploring} utilize a distillation-like mechanism to learn representations by computing the similarity between positives, without the need for negative samples. 
Dimension-based contrastive methods learn visual representations by optimizing the information content of the learned representations and reducing feature redundancy. Barlow Twins~\cite{zbontar2021barlow} endeavors to make the normalized cross-correlation matrix of the augmented embeddings close to the identity matrix. The loss function of VICReg~\cite{bardes2022vicreg} consists of three items: invariance, variance and covariance regularization item. TCR~\cite{li2022neural} employs the Maximum Coding Rate Reduction (MCR$^2$) objective to learn feature subspaces that are both informative and discriminative.
Liu~\cite{liu2022self} proposed using maximum entropy coding for contrastive learning, based on the principle of maximum entropy in information theory, and established a connection between sample-based and dimension-based SSL.

\begin{figure*}[t]
    \centering
    \includegraphics[width=0.96\linewidth]{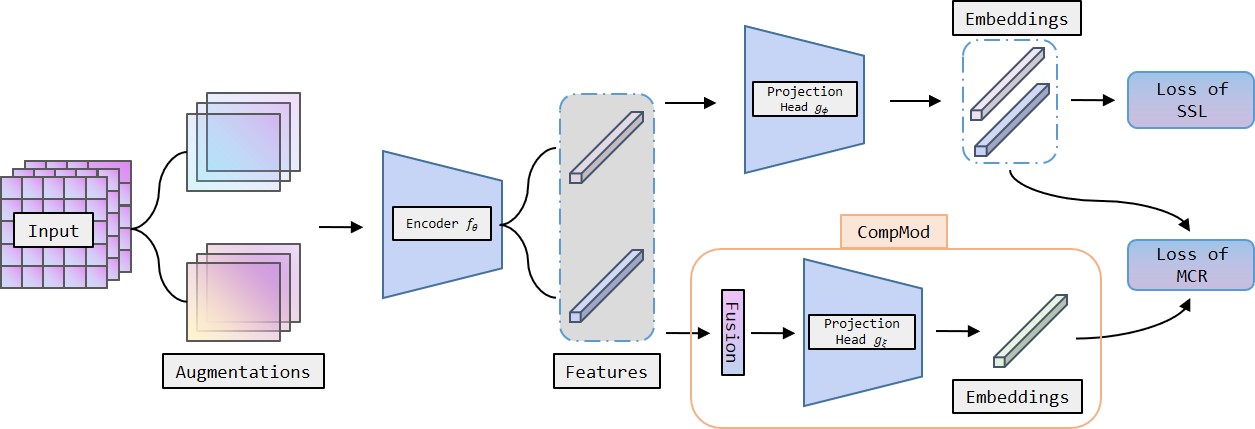}
    \caption{Illustration of self-supervised representation learning framework with Meta Comprehensive Regularization.}
\label{fig:frame}
\end{figure*}

These works mentioned above are based on the invariance of semantic among augmented views, while ignoring the partial loss of label-related information in each view after augmentation, leading to imperfect consistency in semantic information across views. By leveraging the comprehensive information between views, our work allows the feature extractor to  gather more abundant information, thereby inducing the learned sample representations to be more generalizable. Our proposed Meta Comprehensive Regularization can be integrated into existing SSL framework.

\section{Methodology}
Figure~\ref{fig:frame} shows the overview of our proposed method. We design a new module, CompMod, to improve existing self-supervised method. Next, we first theoretically analyze the lack of partial semantic information caused by data augmentation is not conducive to downstream tasks in SSL from an information-theoretic perspective, and then introduce our proposed method and the training process of the model.
\subsection{Contrastive Learning}
Let $D=\{x_i\}_{i=1}^n$ denote the unlabeled training set, where $x_i$ is an input image. Two augmented views $x_i^1$ and $x_i^2$ of the sample $x_i$ are generated by different augmentation strategies $t^1$ and $t^2$ sampled from a augmentation distribution ${A}$. The augmented views are fed into a shared encoder $f_{\theta}$ to obtain their representations $h_i^1=f_{\theta}(x_i^1)$ and $h_i^2=f_{\theta}(x_i^2)$, which are then mapped via a projector $g_{\phi}$ onto the embedding space,  $z_i^1=g_{\phi}(h_i^1)$ and $z_i^2=g_{\phi}(h_i^2)$. We denote the embedding matrix of augmentation view $1$ as $Z_1=[z_1^1,...,z_i^1,...,z_n^1]^T\in R^{n\times d}$, where $d$ is the dimension of the embedding space, so does matrix $Z_2$. Represented by SimCLR, the objective function of the Contrastive Learning (CL) employs the Noise Contrastive Estimation (NCE) loss~\cite{gutmann2010noise}:
\begin{equation}
    \label{loss:instance}
    \mathcal{L}_{ssl} = \mathbb{E}_{z_{i}^{1},z_{i}^{2}} [-\log \frac{e^ { \nicefrac { s(z_{i}^{1}, z_{i}^{2}) }{\tau} }}{e^ {  \nicefrac { s(z_{i}^{1}, z_{i}^{2}) }{\tau} } +\sum\nolimits_{z_j}^{}{e^ { \nicefrac { s(z_{i}^{1}, z_{j}) }{\tau} }}} ]
\end{equation}
where $s(\cdot, \cdot)$ denotes the cosine similarity and $\tau$ represents the temperature hyper-parameter, $z_j$ is the negative sample, $z_j \in \mathcal{{Z}}_1 \cup \mathcal{{Z}}_2/\{z_{i}^{1},z_{i}^{2} \}$, $\mathcal{{Z}}_1$ and $\mathcal{{Z}}_2$ represent the sets of augmented views in the embedding space, respectively.

\subsection{Analysis Based on Information Theory}
%
We assume that the original input images inherently encompass all relevant task-related information, e.g., $I(x;T) = H(T)$, where $x \sim D $ is a random variable, $I$ denotes the mutual information, $H$ represents the information entropy, and $T$ refers to a random variable for the downstream task.
 
As evident from Figure \ref{fig:intro_1},
data augmentation on the input sample results in loss of task-relevant information within the data. 
Consequently, we deduce: $I({x_1};T),I({x_2};T) \leq H(T)$, where ${(x_1, x_2)} \sim \{(x_i^1, x_i^2)\} _{i = 1}^n$.
Also, we can obtain: $I({x_1};{x_2};T) \leq H(T)$. A general explanation for CL is to maximize the mutual information between two augmented views\cite{wang2022rethinking}:
\begin{equation}
\label{MI}
    \max \limits_{f,g}\quad I(z_1; z_2)
\end{equation} 
where $z_1$ and $z_2$ are random variables, ${(z_1,z_2)} \sim {(\mathcal{Z}_1,\mathcal{Z}_2)}$. 
Applying Data Processing Inequality \cite{klir1999uncertainty} to the Markov chain $x \to {x_1(x_2)} \to {z_1(z_2)}$, we have:
\begin{equation}\label{qwefh}
\begin{array}{l}
H(x) \ge I({x_1};{x_2})  \ge I({z_1};{z_2})\\
I(x;T) \ge I({x_1};{x_2};T)  \ge I({z_1};{z_2};T) 

\end{array}
\end{equation}

Based on Eq. \ref{MI} and Eq. \ref{qwefh}, we can draw the conclusion that due to the disruption caused by data augmentation to the semantic information of input samples, contrastive learning is constrained to extract only a subset of task-related information. In order to elucidate this conclusion, we begin by providing a definition for comprehensive representation, followed by deriving the following theorem.

\begin{definition}
    \label{def1}
    (Comprehensive representation) For a random variable $z$ defined in encoder space. $z$ is a comprehensive representation if and only if $I(z; T) = H(T)$.
\end{definition}

\begin{theorem}
    \label{the1}
    (Task-Relevant information in representations) In contrastive learning, given a random variable $x$ representing the original sample space, two random variables $x_1$ and $x_2$ characterizing the sample space after augmentation, and two random variable $z_1$ and $z_2$ denoting the augmented samples within the feature space, we have:
    \begin{equation}
    \begin{array}{l}
    H(T) \ge \{ I({x_1};T),I({x_2};T)\}  \ge I({x_1};{x_2};T)\\
    H(T) \ge \{ I({z_1};T),I({z_2};T)\}  \ge I({z_1};{z_2};T)
    \end{array}
    \end{equation}
    
\end{theorem}

From Theorem \ref{the1}, we deduce the following: (1) Data augmentation leads to a reduction in the amount of task-related information present within the data. (2) The task-related information contained in the commonalities between $z_1$ and $z_2$ is individually less than the task-related information within $z_1$ and $z_2$. Obviously, if the representation of the augmented view is close to the comprehensive representation, it is more beneficial to the downstream tasks. Next, we propose to use Meta Comprehensive Regularization to force the augmented representation to be a comprehensive representation.

\subsection{Meta Comprehensive Regularization}
To address the issue of semantic loss resulting from data augmentation, we propose a new module called CompMod, which learns a more comprehensive representation and helps facilitate model learning, as shown in Figure~\ref{fig:frame}. This module can be directly incorporated into the traditional SSL framework and can complement existing SSL methods. Then, we introduce the module CompMod.

Different augmentations of the same sample are typically derived through various data augmentation techniques, empirically implying that each augmentation results in distinct semantic information loss. As a result, a method that yields a comprehensive feature representation without semantic information loss, distinct from the original input data representation, involves integrating features from different perspectives of the same sample. Thus, a more comprehensive representation of $x_i$ can be obtained by the following formula:
\begin{equation}
\label{eq_c}
\hat{h}_i:=h_{i}^{1}\oplus h_{i}^{2}
\end{equation}
where $\oplus$ represents the fusion strategy, such as the concatenation of vectors: $\hat{h}_i=\left[ h_{i}^{1}\,\,h_{i}^{2} \right] \in R^{2d'}$, where $d'$ is the output dimension of a backbone network. $\hat{h}_i$ fuses the semantic information from all augmented views, so as to solve the problem of partial semantic missing. Next, a projection head $g_{\xi}$ parameterized by $\xi$ maps $\hat{h}_i$ onto the same embedding space as $z_i^1$ and $z_i^2$. And then, we obtain the so-called more comprehensive embedding of sample $x_i$, denoted as $\hat{z}_i=g_{\xi}(\hat{h}_i) \in R^d$. The comprehensive embedding matrix is defined as $\hat{Z}=[\hat{z}_1,...,\hat{z}_i,...,\hat{z}_n]^T \in R^{n\times d}$.

Simple fusion alone does not guarantee that the learned features encompass all semantics. Inspired by the maximum entropy principle in information theory, a generalizable representation should be the one with the maximum entropy among all possible representations, corresponding to the maximization of the semantic information associated with the true label. Here, we use the code length of the lossy data coding~\cite{cover1999elements} to calculate the entropy of the embedding matrix $\hat{Z}$, which apply the Taylor series expansion:
\begin{equation}
    \mathcal{L}_{comp}(\hat{Z})= -{\rm{Tr}}( \mu \sum_{k=1}^m{\frac{( -1 ) ^{k+1}}{k}( \lambda {\hat{Z}}{\hat{Z}}^T ) ^k} ) 
\end{equation}
where $k$ is the order of Taylor expansion, $\mu$ and $\lambda$ are hyperparameters. Therefore, to further ensure the comprehensiveness of the obtained $\hat{Z}$, we propose that the obtained $\hat{Z}$ should minimize $\mathcal{L}_{comp}(\hat{Z})$.

Next, we use the comprehensive representation to guide the learning of the backbone network. To enhance the semantic richness of the representation in each augmented view, we constrain the information contained in the augmented embeddings $Z_1, Z_2$ to equal the information contained in the comprehensive embedding $\hat{Z}$. We propose to minimize the following loss:
\begin{equation}
\label{loss_mc}
    \mathcal{L}_{mcr}(Z_1,Z_2) =-\sum_{t=1}^2{{\rm{Tr}}( \mu {\sum_{k=1}^m{\frac{( -1 ) ^{k+1}}{k}( \lambda \hat{Z}Z_{t}^{T} ) ^k}} )}
\end{equation}
where $Z_{t}$ is the embedding matrix of augmented view, $t=1,2$. As we can see, when $\hat{Z}$ is predetermined and carries maximal information content, in order to minimize Eq. \ref{loss_mc}, it is necessary for $Z_{t}$ to be equal to $\hat{Z}$. Thus, minimizing Eq. \ref{loss_mc} can be considered as a conduit for transferring comprehensive information from $\hat{Z}$ to both $Z_{1}$ and $Z_{2}$, enabling them to compensate for the semantic loss incurred by data augmentation. Consequently, while extracting consistent semantic information from $Z_{1}$ and $Z_{2}$, minimizing Eq. \ref{loss_mc} facilitates the extraction of comprehensive semantic information.

\subsection{Model Objective}

Finally, we present the objective during the training phase, which can be divided into two steps. The first step is to learn $f_{\theta}$ and $g_{\phi}$ that can extract feature representation. The second step is to learn $g_{\xi}$ that can obtain comprehensive representation by a bi-level optimization mechanism. The training process is shown in Algorithm \ref{alg1}.

Specifically, in the first step of each epoch, we fix $g_{\xi}$ and update $f_{\theta}$ and $g_{\phi}$ through the following formulation:
\begin{equation}\label{dasd}
    \{ \theta ,\phi \} = \{ \theta ,\phi \} - 
    r \cdot \nabla _{ \theta ,\phi }( \mathcal{L}_{ssl} + \lambda_1 \mathcal{L}_{mcr})
\end{equation}
where $r$ is the learning rate and $\lambda_1$ is the hyperparameter. The purpose of learning Eq. \ref{dasd} is to extract consistency semantic information between $Z_1$ and $Z_2$. However, $\mathcal{L}_{ssl} + \lambda_1\mathcal{L}_{mcr}$ in Eq. \ref{dasd} enables both $Z_1$ and $Z_2$ to simultaneously possess comprehensive semantic information. Thus, learning Eq. \ref{dasd} can result in the consistency information between $Z_1$ and $Z_2$ being comprehensive information.

\begin{algorithm}[tb]
\caption{The main algorithm}
\label{alg1}
\textbf{Input}: Training set $D$; Batch Size $n$; Encoder function $f_\theta$; Projection Head $g_\phi$; Multi-layer Network  $g_\xi$.\\
\textbf{Parameter}: Regularization Parameter: $\lambda_1, \lambda_2$\\
\textbf{Output}: The optimal encoder: $f_{\theta ^*}$
\begin{algorithmic}[1] 
\FOR{sample batch $X$ from $D$}
        \STATE \# generate two augmented views
	\STATE $x_i^1,x_i^2 = t_1(x_i), t_2(x_i)$, $t_1, t_2\in A, x_i \in X$
	\STATE \# obtain the augmented embeddings\\
	\STATE $z_i^1 = g_\phi(f_\theta(x_i^1))$\\
	\STATE $z_i^2 = g_\phi(f_\theta(x_i^2))$\\
	\STATE \# obtain the more comprehensive embedding\\
	\STATE $\hat{z}_i = g_\xi(\hat{h}_i), where\ \hat{h}_i = h_{i}^{1}\oplus h_{i}^{2}$\\
        \STATE \# Under the fixed $\xi$, update $\{\theta_{\xi},\phi_{\xi}\}$\\
        \STATE $\{\theta_{\xi},\phi_{\xi}\}   = \{\theta ,\phi \}   - r \cdot {\nabla _{\theta ,\phi }}({{\cal L}_{ssl}} + {\lambda _1}{{\cal L}_{mcr}})$

        \STATE \# Under the fixed $\{\theta_{\xi},\phi_{\xi}\}$, update $\xi$\\
        \STATE $\xi  = \xi  - r \cdot {\nabla _\xi }({\mathcal{L}_{ssl}}({\theta }_{\xi},{\phi }_{\xi}) + {\lambda _2} {\mathcal{L}_{comp}})$
\ENDFOR
\end{algorithmic}
\end{algorithm}

In the second step of each epoch, we fix $f_{\theta}$ and $g_{\phi}$ and update $g_{\xi}$ through the following formulation:
\begin{equation}\label{casd}
\begin{array}{l}
\xi  = \xi  - r \cdot {\nabla _\xi }({\mathcal{L}_{ssl}}({\theta }_{\xi},{\phi }_{\xi}) + {\lambda _2} {\mathcal{L}_{comp}})\\
s.t. \{\theta_{\xi},\phi_{\xi}\}   = \{\theta ,\phi \}   - r \cdot {\nabla _{\theta ,\phi }}({{\cal L}_{ssl}} + {\lambda _1}{{\cal L}_{mcr}})\\
\end{array}
\end{equation}
where $\mathcal{L}_{ssl}({\theta }_{\xi},{\phi }_{\xi})$ represents that the loss $\mathcal{L}_{ssl}$ is calculated based on $f_{\theta_{\xi}}$ and $g_{\phi_{\xi}}$, and $\lambda_2$ is the hyperparameter. It is important to note that during the computation of $\mathcal{L}_{ssl}$, $g_{\xi}$ is not involved, hence direct differentiation of $\mathcal{L}_{ssl}$ with respect to ${\xi}$ is not possible. However, when computing matrix $\mathcal{L}_{ssl}({\theta }_{\xi},{\phi }_{\xi})$, as indicated by Eq. \ref{casd}, $\theta_{\xi}$ and $\phi_{\xi}$ can be treated as functions of $\xi$, allowing for direct differentiation of $\mathcal{L}_{ssl}({\theta }_{\xi},{\phi }_{\xi})$ with respect to $\xi$. Simultaneously, optimizing ${\nabla _\xi }{\mathcal{L}_{ssl}}({\theta }_{\xi},{\phi }_{\xi})$ can be conceptualized as follows: by manipulating $\xi$ to induce changes in ${\mathcal{L}_{ssl}}({\theta }_{\xi},{\phi }_{\xi})$, constrained by the conditions outlined in Eq. \ref{casd}, where ${\mathcal{L}_{ssl}}({\theta }_{\xi},{\phi }_{\xi})$ is consistently optimized under these $\xi$-induced circumstances. Subsequently, among all optimal states of ${\mathcal{L}_{ssl}}({\theta }_{\xi},{\phi }_{\xi})$, the objective is to identify a configuration that minimize the magnitude of ${\nabla _\xi }{\mathcal{L}_{ssl}}({\theta }_{\xi},{\phi }_{\xi})$. So, optimizing ${\nabla _\xi }{\mathcal{L}_{ssl}}({\theta }_{\xi},{\phi }_{\xi})$ can be simply regarded as the mechanism for modulating $\xi$ to reduce its magnitude, thereby enhancing the similarity between matrices $Z_1$ and $Z_2$. As widely acknowledged, it is only when matrices $Z_1$ and $Z_2$ encapsulate a greater abundance of analogous semantic information that matrices $Z_1$ and $Z_2$ can exhibit increased similarity. Therefore, optimizing ${\nabla _\xi }{\mathcal{L}_{ssl}}({\theta }_{\xi},{\phi }_{\xi})$ can compel $g_{\xi}$ to extract a greater amount of semantic information. When coupled with the influence of $\mathcal{L}_{comp}$, this process enables Eq. \ref{casd} to coerce $g_{\xi}$ into capturing comprehensive semantic information.

\subsection{Causal Interpretation}
\begin{figure}[t]
    \centering
    \includegraphics[width=0.92\linewidth]{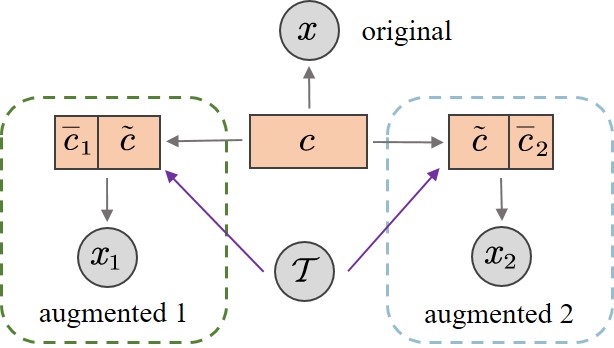}
    \caption{Structural causal model of latent variables. We assume that part of the semantic missing in the augmented view $x_1$ and $x_2$ compared to the original view $x$. 
    These exclusive semantic $\bar{c}$ is affected by the semantic $c$ and the applied augmentation strategy $t$.}
    \label{fig:SCM}
\end{figure}

We use a Structural Causal Model (SCM)~\cite{pearl2018book} to describe the causal relationship between variables in self-supervised learning, whereby data augmentation can be viewed as a counterfactual. We define $c$ to be the semantic of original sample, $\tilde{c}$ is the invariant semantics across augmented views, $\bar{c}$ is the exclusive semantics in the augmented view. We assume that the samples are only generated by their semantic information (or label) and exogenous variables, as shown in Fig.\ref{fig:SCM}. The augmentation strategy $t$ is randomly sampled from the strategy set $\mathcal{T}$. 
We can formulate the relation between these variables:
\begin{equation}
\nonumber
\begin{aligned}
     c := f_c(u_c), \tilde{c} := f_{\tilde{c}}(c, t), \bar{c} := f_{\bar{c}}(c, t),
     x:=f(c, u_x);
\end{aligned}
\end{equation}
where $u_c, u_x$ is independent exogenous variable, and $f_{\tilde{c}}, f_{\bar{c}}$ are deterministic functions. 

Given a factual observation $x' = f(c', u_{x'})$. If a new augmentation strategy $t^{*}$ is applied, it is equivalent to an intervention mechanism affecting $f_{\tilde{c}}, f_{\bar{c}}$:
\begin{equation}
\nonumber
    do(\tilde{c} := f_{\tilde{c}}(c', t^{*}), \bar{c} := f_{\bar{c}}(c', t^{*}))
\end{equation}
Using the modified SCM by fixing all exogenous variables, the relation between the variables in the counterfactual sample $x''= f(c', u_{x''})$ and its augmented views can be reformulated as:
\begin{equation}
\nonumber
    c'' := c', \tilde{c}' := f_{\tilde{c}}({c', t^{*}}), \bar{c}' := f_{\bar{c}}({c', t^{*}});
\end{equation}
Thus, data augmentation in self-supervised learning can be viewed as counterfactual. Then, we can obtain:
\begin{theorem} 
    Assume that the data generating
process is consistent with the above description. Let $f:x\rightarrow z$ be any smooth function that can minimize the following objective:
    \begin{equation}\label{cfghkm}
         E_{\left( z^1_{i},z_{i}^2 \right) \sim \left( \mathcal{Z}_1,\mathcal{Z}_2 \right)}\left[ \left\| z_{i}^1 - z_{i}^2 \right\| _{2}^{2} \right] - H_{min}(z_{i}^t)
    \end{equation}
    where $H_{min}(\cdot)$ denotes the minimum entropy of all augmented embeddings, $z_i^t = \{z_i^1, z_i^2\}$. In this way, the learned encoder $f_{\theta}$ can capture all semantic information related to the original sample.
\end{theorem}

\begin{table*}[ht]
    \centering
    \begin{tabular}{l|cc|cc|cc|cc}
    \toprule
    \multirow{2}*{Methods} & \multicolumn{2}{c|}{CIFAR-10} & \multicolumn{2}{c|}{CIFAR-100} & \multicolumn{2}{c|}{STL-10} & \multicolumn{2}{c}{Tiny ImageNet}
    \\
    ~ & linear & 5-nn & linear & 5-nn & linear & 5-nn & linear & 5-nn \\
    \hline
    SimCLR &  91.80  & 88.42 & 66.83  & 56.56 & 90.51  & 85.68 & 48.82  & 32.86 \\
    BarlowTwins &  90.88  & 88.78 & 66.67  & 56.39 & 90.71  & 85.31 & 49.74  & 33.61 \\
    BYOL &  91.73  & 89.45 & 66.60  & 56.82 & 91.99  & 88.64 & 51.00  & 36.24 \\
    SimSiam&  91.51  & 89.31 & 66.73  & 56.87 & 91.92  & 88.54 & 50.92  & 35.98 \\
    W-MSE &  91.99  & 89.87 & 67.64  & 56.45 & 91.75  & 88.59 & 49.22  & 35.44 \\
    SwAV &  90.17  & 86.45 & 65.23  & 54.77 & 89.12  & 84.12 & 47.13  & 31.07 \\
    SSL-HSIC &  91.95  & 89.91 & 67.22  & 57.01 & 92.06  & 88.87 & 51.42  & 36.03 \\
    VICReg &  91.08  & 88.93 & 67.15  & 56.47 & 91.11  & 86.24 & 50.17  & 34.24 \\
    \hline
    SimCLR+ &  93.91  & 91.21 & \textbf{68.91}  & \textbf{58.22} & 92.77  & 87.98 & 51.01  & 35.23 \\
    BYOL+ &  \textbf{93.96}  & \textbf{91.53} & 68.74  & 58.01 & \textbf{94.95}  & \textbf{89.88} & \textbf{53.51}  & \textbf{37.95} \\
    BarlowTwins+ &  92.54  & 90.75 & 68.29  & 57.84 & 93.12  & 89.61 & 51.72  & 35.21 \\
    \hline
    \end{tabular}
    \caption{Classification accuracy on small and medium datasets. Top 1 accuracy(\%) of linear classifier and a 5-nearest neighbors classifier for different datasets with a ResNet-18. Best results are in bold.}
    \label{tab:exp1_1}
\end{table*}

The above theorem states that when we constrain the embeddings to have the maximum entropy, the SSL model can constrain the learned representation to contain more semantics. Note that the Eq. \ref{cfghkm} is similar to Eq. \ref{casd}. Therefore, this theorem also provides a theoretical basis for the proposed bi-level learning method.

\section{Experimental Results}
\label{exp}

In this section, we first evaluate it on classification task using linear evaluation and semi-supervised learning settings. Then, we validate our method on object detection and instance segmentation tasks in computer vision.

\subsection{Experimental Setting}

\textbf{Datasets.}
For the classification task, we evaluate our proposed method on the following six image datasets, including CIFAR-10 and CIFAR-100 dataset~\cite{krizhevsky2009learning}, STL-10 dataset ~\cite{coates2011analysis}, Tiny ImageNet dataset~\cite{le2015tiny}, ImageNet-100 dataset~\cite{russakovsky2015imagenet}, and ImageNet dataset~\cite{russakovsky2015imagenet}. 
For transfer learning, we validate our method by the performance on the object detection and semantic  segmentation tasks on COCO~\cite{lin2014microsoft} dataset.

\textbf{Default Setting.}
Each input sample generates two corresponding positive samples in the experiment. The image augmentation strategies comprise the following image transformations: random cropping, resizing, horizontal flipping, color jittering, converting to grayscale and gaussian blurring. Detailed experimental settings for different downstream tasks can be found in Appendix B. In the experiment, we use Resnet18 or Resnet50 as our base encoder network, along with a 3-layer MLP projection head to project the representation to a embedding space.

\textbf{CompMod Details.} The CompMod consists of a multi-layer linear network, which is set to $2d'-d'-d$, where $d'$ is the output dimension of the backbone network $f_\theta$.

\subsection{Downstream Tasks}

\subsubsection{Self-supervised Learning}
We conduct a classification task to test our proposed method. For comparison,  we take SimCLR~\cite{chen2020simple}, BarlowTwins~\cite{zbontar2021barlow}, BYOL~\cite{grill2020bootstrap}, SimSiam~\cite{chen2021exploring}, W-MES~\cite{ermolov2021whitening}, SwAV~\cite{caron2020unsupervised}, MoCo~\cite{he2020momentum}, CMC~\cite{tian2020contrastive}, SSL-HSIC~\cite{li2021self} and VICReg~\cite{bardes2022vicreg} as baselines.
\begin{table}[t]
    \centering
    \begin{tabular}{l|cc|cc}
		\toprule
		\multirow{2}*{Methods} & \multicolumn{2}{c|}{ImageNet-100} & \multicolumn{2}{c}{ImageNet}
		\\
		~ & top-1 & top-5 & top-1 & top-5 \\
		\hline
		SimCLR &  70.15  & 89.75 & 69.32  & 89.15\\
		MoCo &  72.81  & 91.64 & 71.13  & -\\
		CMC & 73.58 & 92.06 & 66.21 & 87.03 \\
		BYOL &  74.89  & 92.83 & 74.31  & 91.62\\
		SwAV &  75.77  & 92.86 & 75.30  & -\\
		DCL & 74.60 & 92.08 & - & - \\
		RELIC & - & - &  74.81 & 92.23 \\
		SSL-HSIC &  -  & - & 72.13  & 90.33\\
		ICL-MSR &  72.08  & 91.60 & 70.73  & 90.43\\
		BarlowTwins & 72.88 & 90.99 & 73.22 & 91.01\\
		\hline
		SimCLR+ &  72.21 & 91.23 & 71.89  & 91.52\\
		BYOL+ &  \textbf{76.95}  & 93.94 & 75.11  & \textbf{93.55}\\
		BarlowTwins+ &  76.88  & \textbf{94.11} & \textbf{75.62}  & 92.13\\
		\hline
    \end{tabular}
    \caption{Evaluation on ImageNet-100 and ImageNet datasets. The representations are obtained with a ResNet-18 with our method on top 1 accuracy(\%) of linear classifier and a 5-nn classifier. Best results are in bold.}
    \label{tab:exp1_2}
\end{table}
We validate the proposed method with the results of a linear classifier and a 5-nearest neighbor classifier. Table \ref{tab:exp1_1} shows the performance of different SSL methods, where ``method+'' denotes our proposed method. The results show that our proposed method improves the classification performance, in which SimCLR+ and BYOL+ improve by more than 2\% on CIFAR10 and CIFAR100 dataset, while BYOL+ improves by about 2.5\% on Tiny ImageNet dataset.

Furthermore, we test our method for classification on two larger datasets, ImageNet-100 and ImageNet. For comparison, we also add several other methods including MoCo~\cite{he2020momentum}, CMC~\cite{tian2020contrastive}, ICL-MSR~\cite{qiang2022interventional} and RELIC~\cite{mitrovic2020representation}. The results in Table \ref{tab:exp1_2} demonstrate that our method still improves over the baseline, e.g., BarlowTwins+ achieves 4\% performance improvement on ImageNet-100, SimCLR+ and BarlowTwins+ achieve more than 2\% on ImageNet.

\subsubsection{Semi-supervised Learning}
The detailed experimental setup follows the most common
evaluation protocol for semi-supervised learning, as in Appendix B. Table \ref{tab:semi} reports the classification results on ImageNet compared with existing methods using two pre-trained models. From the results, Barlow Twins+ is 1.1\% better than Barlow Twins, and BYOL+ increases by about 1.4\% at the 1\% subset setting.

\begin{table}[t]
    \centering
    \setlength{\tabcolsep}{1.6mm}
    \begin{tabular}{lccccc}
		\toprule
		\multirow{2}*{Methods} & \multirow{2}*{Epochs}  & \multicolumn{2}{c}{1\%} & \multicolumn{2}{c}{10\%} \\\cline{3-6}
		~ & ~ & top-1 & top-5 & top-1 & top-5 \\
		\hline
		SimCLR & 1000 &  48.3  & 75.5 & 65.6  & 87.8\\
		BYOL & 1000 & 53.2  & 78.4 & 68.8 & 89.0\\
		SwAV & 1000 & 53.9  & 78.5 & \textbf{70.2}  & \textbf{89.9}\\
		BarlowTwins & 1000 & 55.0  & 79.2 & 69.7  & 89.3\\
		\hline
		SimCLR+ & 1000 & 49.1 &  75.8 & 65.8  & 88.0\\
		BYOL+ & 1000 & 54.6  & 78.9 & 69.2  & 89.3\\
		Barlow Twins+ & 1000 & \textbf{56.1}  & \textbf{79.8} & \textbf{70.2} & \textbf{89.9}\\
		\hline
    \end{tabular}
\caption{Semi-supervised classification. We finetune the pre-trained model using 1\% and 10\% training samples of ImageNet following ~\cite{zbontar2021barlow}, and the top-1 and top-5 under linear evaluation are reported.}
\label{tab:semi}
\end{table}

\begin{table*}[ht]
\centering
    \begin{tabular}{c|ccccc|ccc}
		\toprule
		\multirow{2}*{ID}& \multicolumn{5}{c|}{Data augmentations} & \multicolumn{3}{c}{Methods} \\
		\cline{2-9}
        & \parbox[c]{1.4cm}{horizontal flip} & {rotate} & \parbox[c]{1.2cm}{random crop} & \parbox[c]{1.2cm}{random grey} & \parbox[r]{0.8cm}{color\quad jitter} & SimCLR+ & BYOL+ & Barlow Twins+ \\
		\hline
		1 & $\checkmark$ & $\checkmark$ & & & & 93.65 & 92.64 & 91.75 \\
		2 & & & $\checkmark$ & & & 92.31 & 92.78 & 92.09 \\
		3 & & & & $\checkmark$ & & 92.78 & 93.16 & 92.15 \\
		4 & & & & & $\checkmark$ & 93.36 & 92.99 & 91.95 \\
		5 & $\checkmark$ & & $\checkmark$ & & & 93.47 & 92.74 & 92.23 \\
		6 & & $\checkmark$ & & & $\checkmark$ & 93.72 & 92.89 & 91.89 \\
		7 & $\checkmark$ & & $\checkmark$ & $\checkmark$ &  & 93.75 & 93.78 & 92.48 \\
		8 & $\checkmark$ & $\checkmark$ & $\checkmark$ & $\checkmark$ & $\checkmark$  & \textbf{93.91} & \textbf{93.96} & \textbf{92.54} \\
	\hline
    \end{tabular}
    \caption{Comparison of different data augmentations by using a ResNet-18 on CIFAR-10 dataset.}
\label{tab:distur}
\end{table*}

\subsubsection{Transfer Learning}
We evaluate our method for the localization based tasks of object detection and instance segmentation on COCO~\cite{lin2014microsoft} datasets. ImageNet supervised pre-training is often used as initialization for fine-tuning downstream tasks. Several different self-supervised methods are used for performance comparison.  We report the results of our proposed method compared with baselines in Table \ref{tab:transfor}, showing that the proposed method brings performance improvements on different downstream tasks.

\begin{table}[t]
\centering
    \setlength{\tabcolsep}{1.6mm}
    \begin{tabular}{lcccccc}
		\toprule
		\multirow{2}*{Methods} & \multicolumn{3}{c}{Object Det.} & \multicolumn{3}{c}{Instance Seg.} \\
		\cline{2-4}
		\cline{5-7}
		~ & AP & AP$_{50}$ & AP$_{75}$ & AP & AP$_{50}$ & AP$_{75}$ \\
		\hline
		Supervised & 38.2 &  58.2  & 41.2 & 33.3  & 54.7 & 35.2 \\
        \hline
		SimCLR & 37.9 &  57.7  & 40.9 & 33.2  & 54.6 & 35.3 \\
		SwAV & 37.6 &  57.6  & 40.2 & 33.0  & 54.2 & 35.1 \\
		BYOL & 37.9 &  57.8  & 40.9 & 33.1  & 54.3 & 35.0 \\
		SimSiam & 37.9 &  57.5  & 40.9 & 33.3  & 54.2 & 35.2 \\
		BarlowTwins & 39.2 & 59.0  & 42.5 & 34.2  & 56.0 & 36.5 \\
		\hline
		SimCLR+ & 38.1 & 58.1 & 41.0 & 33.7 & 54.6 & 35.1\\
		BYOL+ & \textbf{39.7} & 59.1  & 42.9 & \textbf{35.4} & 56.1 & 36.2\\
		Barlow Twins+ & 39.1 & \textbf{59.3} & \textbf{43.1} & 35.2 & \textbf{56.2} & \textbf{36.9}\\
		\bottomrule
    \end{tabular}
    \caption{Transfer learning. We pre-train the network on ImageNet dataset. Then,  we learn representation on the object detection and instance segmentation tasks on COCO dataset using Mask P-CNN. Evaluation is on AP, AP$_{50}$ and AP$_{75}$.}
\label{tab:transfor}
\end{table}

\subsection{Ablation Experiments}
\subsubsection{Parametric Sensitivity}
In this section, we conduct an experimental investigation of the model trade-off parameters. Specifically, we vary $\lambda_1$ and $\lambda_1$ in the range of [0.001, 0.01, 0.1, 1], and record the classification accuracy of our method using a ResNet-18 on CIFAR-10 dataset with the SimCLR+ method. The results in Table \ref{tab:para} indicates that our method has minimal variation in accuracy, indicating that hyperparameter tuning is easy in practice.

\subsubsection{Analysis of Data Augmentation}
we compare the linear classification accuracy under different augmentation strategies on CIFAR-10 dataset as shown in Table \ref{tab:distur}. As can be seen, there is no significant difference in classification accuracy, indicating that our method can be applied to different augmentation strategies.

\subsubsection{Fusion Strategy and Optimization}

In this section, we first investigate the impact of different fusion strategies.

Mixup ~\cite{verma2021towards} can be used to fuse features in representation space. We can obtain the more comprehensive representation using the following formula:
\begin{equation}
	\label{eq_mixup}
	\hat{h}_i= \alpha * h_{i}^{1} + (1 - \alpha) * h_{i}^{2}
\end{equation}
where $\alpha$ is a coefficient sampled from a uniform distribution, $\alpha \sim U(0,1)$. By adjusting $\alpha$, we can control the semantic information to be biased towards $h_{i}^{1}$ or $h_{i}^{2}$.  
Another strategy is to achieve semantic fusion in the embedding space:
\begin{equation}
	\tilde{z}_i := z_{i}^{1}\oplus z_{i}^{2}=\left[ z_{i}^{1}\,\,z_{i}^{2} \right]
\end{equation}
Then $\tilde{z}_i$ is mapped onto the embedding space to obtain $\hat{z}_i$ via a projection head $g_\zeta$ composed of a multi-layer linear network $(2d-d-d)$: $\hat{z}_i=g_\zeta(\tilde{z}_i)$.

\begin{table}[t]
\centering
    \begin{tabular}{ccccc}
	\toprule
        \diagbox{$\lambda_2$}{$\lambda_1$}  & 0.001 & 0.01 & 0.1 & 1 \\
	\hline
        0.001 & 91.79 & 92.13 & 92.77 & 91.75\\
        0.01 & 92.56 & 91.89 & 93.91 & 91.11\\
        0.1 & 93.35 & 92.23 & 92.35 & 90.78\\
        1 & 93.03 & 91.26 & 92.77 & 90.08\\
	\hline
    \end{tabular}
    \caption{Parametric analysis of $\lambda_1$ and $\lambda_2$.}
\label{tab:para}
\end{table}

\begin{table}[t]
\centering
    \setlength{\tabcolsep}{1.4mm}
    \begin{tabular}{l|ccccc|c|c}
		\toprule
         & \multicolumn{5}{c|}{Mixup} & \makebox[0.02\textwidth]{M($h$)} & \makebox[0.02\textwidth]{M($z$)} \\
         \hline
        $\alpha$ & \makebox[0.001\textwidth]{0.1} & \makebox[0.01\textwidth]{0.3} & \makebox[0.01\textwidth]{0.5} & \makebox[0.01\textwidth]{0.7} & \makebox[0.01\textwidth]{0.9} & - & - \\
        \hline
        Acc. & 91.03 & 91.33 & 91.87 & 91.34 & 91.56 & \textbf{93.96} & 92.58 \\
        \hline
        \multicolumn{6}{l|}{No bi-level} & \multicolumn{2}{c}{92.05} \\
    \hline
    \end{tabular}
    \caption{Analysis of Fusion Strategy and Optimization. Experimental results are based on the classification with BYOL+. M($h$) means fusion in representation space, while, M($z$) means fusion in embedding space}
\label{tab:fusion}
\end{table}

Additionally, our proposed method utilizes a bi-level
optimization mechanism for model optimization during training. In this section, we also analyze the impact of not using this strategy in experiments.

Table \ref{tab:fusion} shows the results of BYOL+ utilizing different feature fusion strategies and optimization on the classification task on CIFAR-10 dataset. Experimental results demonstrate that the fusion strategy and optimization we adopt achieve the best results.

\section{Conclusion}
In this paper, we find that data augmentation in SSL may lead to the lack of task-related information from information theory,resulting in a reduction of the model's performance in downstream tasks.  
To this end, we design a novel module CompMod with Meta Comprehensive Regularization as a complement to existing SSL frameworks. CompMod exploits a bi-level optimization mechanism and constraint based on maximum entropy coding to enable more information to be discovered, thereby enhancing the generalization of the learned model.
Moreover, a causal interpretation provide theoretical support for the proposed method. Finally, the performance of various downstream tasks validates the effectiveness of our proposed method.

\bibliography{aaai24}

\end{document}